\documentclass[runningheads]{llncs}

\usepackage[final,year=2026]{eccv}
\usepackage[width=122mm,left=12mm,paperwidth=146mm,height=193mm,top=12mm,paperheight=217mm]{geometry}

\usepackage{eccvabbrv}

\usepackage{graphicx}
\usepackage{float}
\usepackage{booktabs}
\usepackage{amsmath}
\usepackage{amssymb}
\usepackage[accsupp]{axessibility}

\usepackage{hyperref}
\usepackage{orcidlink}
\usepackage{caption}

\begin{document}

\title{Embeddings based Anomaly Detection for Cleaning Global 
Crop Type 
Reference Datasets}
\titlerunning{Embedding-based Outlier Detection for Crop Reference Data}

\author{Syed Roshaan Ali Shah\inst{1}$^{\star}$\orcidlink{0000-0001-9517-5142} \and
Kristof Van Tricht\inst{1}$^{\star}$\orcidlink{0000-0001-5727-9098} \and
Christina Butsko\inst{1}\orcidlink{0009-0001-0975-5413} \and
Jeroen Degerickx\inst{1}\orcidlink{0000-0001-9022-9623} \and
Zoltan Szantoi\inst{2}\orcidlink{0000-0003-2580-4382}}
\authorrunning{Embedding-based Outlier Detection for Crop Reference Data}
\institute{VITO Remote Sensing, Mol, Belgium \and
European Space Agency (ESA), Frascati, Italy}

\maketitle
\let\svthefootnote\thefootnote
\let\thefootnote\relax\footnotetext{$^{\star}$~Equal contribution}
\let\thefootnote\svthefootnote

\begin{abstract}
High quality reference data remain a critical bottleneck for crop-type mapping at any spatial and temporal scale. Operational systems such as WorldCereal aggregate labels from heterogeneous sources such as parcel registers, national databases, field surveys, and map-derived products, each with their own biases, coverage gaps and unknown label noise. Simple global rules are inadequate, since crop phenology and observation conditions vary strongly across regions and seasons. In this study, we focus on a single, operationally relevant question: whether embeddings produced through geospatial foundation models are a viable basis for cleaning the reference data. We propose a practical, locality-aware, embedding-based anomaly (EBA) detection framework that operates on the embeddings of a pretrained Earth-observation encoder. We score each labelled sample against other samples of the same crop in the same area using a pretrained embedding, flag the ones that stand out, and test whether removing or down-weighting them before training yields a better model. Two complementary tests indicate that the flagged points behave like genuine label errors rather than ordinary samples: against synthetic ground truth, the detector concentrates injected label errors $2.5$--$5\times$ above chance in its flagged set (detection AUROC up to $0.84$); and on real data, a model-independent test shows that removing or confidence-weighting the flagged held-out points raises the measured accuracy of already-trained models, for both crop type and land cover. Acting on the flags then improves the WorldCereal crop-type model across five macro-regions, evaluated on a fixed held-out split under three views. We find conservative cleaning helps while over-cleaning hurts. The EBA detector approach is designed to be reproducible and extensible, and can serve as a template for cleaning large, noisy Earth observation reference datasets beyond crop mapping.
 The reference code is available through the \href{https://github.com/WorldCereal/EBA_detector}{GitHub repository}.
\end{abstract}

\section{Introduction}
\label{sec:intro}

Supervised and semi-supervised crop-type mapping at global scale depends critically on access to harmonised, reliable reference data. In practice such labels are assembled from diverse sources: parcel-level registers such as LPIS in Europe, national databases, field surveys like LUCAS~\cite{dandrimont2020lucas}, contractor datasets, and increasingly map-derived labels from national or global products such as the USDA Cropland Data Layer~\cite{boryan2011cdl}. Each source carries its own spatial footprint, class definitions, temporal coverage and error modes. Combined into a single training pool, the resulting global dataset exhibits unknown label noise, regional inconsistencies and artefacts from the underlying maps and sampling strategies, a recurring obstacle in deploying global crop-mapping models~\cite{butsko2025lessons}. Label noise is known to degrade supervised learning and is the subject of a large literature on its characterisation and mitigation~\cite{frenay2014survey,song2022labelnoise,northcutt2021confident,northcutt2021pervasive}; it is especially damaging for land-cover and crop mapping from satellite image time series~\cite{pelletier2017noise,elmes2020accounting}.

The notion of an outlier in Earth-observation (EO) data is itself dual. Some outliers are genuine extreme events such as droughts, floods or heatwaves that are environmentally real and may be scientifically valuable~\cite{flach2017multivariate}; others are label errors or artefacts that should be cleaned before training. Our focus is the latter: identifying samples whose representation is highly anomalous relative to other samples of the same declared class in the same locality, a strong indicator of mislabelling rather than of an interesting extreme event.

WorldCereal~\cite{vantricht2023WorldCereal,see2023dynamic} is an open, modular system developed under ESA funding to provide global-scale products. It combines a reference-data module, ingestion and harmonisation pipelines, multi-modal feature extraction and learning components to deliver crop extent, crop-type and irrigation maps at seasonal scales. Its Reference Data Module (RDM)~\cite{boogaard2023rdm} hosts harmonised reference datasets, exposes flexible querying and download, and lets external contributors upload and share samples. The RDM is thus the central entry point for training data, but at the same time concentrates the challenge of combining heterogeneous, noisy labels.

Two characteristics make anomaly detection in this setting non-trivial. First, there is no global, invariant signature for any crop that would hold across climates, management practices and sensor conditions: what a normal example looks like for winter wheat in temperate Europe differs entirely from irrigated rice in the tropics or mixed-cropping systems in semi-arid regions. Second, modern EO pipelines increasingly summarise rich, multi-modal time series with self-supervised representation models~\cite{tseng2023presto,cong2022satmae,russwurm2020selfattention,jakubik2023foundation}, in whose feature space distinct classes may still overlap. Anomaly detection must therefore be local in space and label domain, and robust to high intra-class variability~\cite{pang2021deep}. Many such errors are invisible in a basemap or RGB chip and surface only in the multi-temporal, multi-modal signal.

These self-supervised representation model pipelines already produce, for every reference sample, a compact embedding vector that summarises its full multi-modal time series. 
We study whether such embeddings are a viable basis for cleaning the reference data, and we develop a locality-aware anomaly detector that operates directly on them. The method is presented in the context of the WorldCereal crop-type dataset but is designed to generalise to other EO reference collections.

We make three contributions: (i)~a generic \emph{local-slice} formulation of outlier detection for labelled EO reference data, in which each sample is judged only against same-class neighbours in the same locality; (ii)~an embedding-based detector that combines centroid distance and local neighbour structure into an anomaly score with graded categories and a confidence output; and (iii)~an evaluation combining ground-truthed synthetic-noise recovery, a model-independent test on real data, and a downstream study on the operational WorldCereal model across five macro-regions.

Outlier detection here is not the detection of an absolute property of a point but of a point being unusual relative to a chosen reference set; that set is therefore constructed explicitly as the local slice, and all scoring is relative to it.

\section{Data and Methods}

\subsection{Reference data}
We work with labelled reference data of the kind hosted in repositories such as the WorldCereal Reference Data Module (RDM)~\cite{boogaard2023rdm}, which can serve crop mapping at any scale. Each sample is a point, typically a field centroid, with a geographic coordinate and a land-cover and crop-type label. For crop-mapping analyses these samples are paired with co-located multi-temporal Earth-observation signals, typically optical (Sentinel-2), radar (Sentinel-1), meteorological and terrain variables, and optionally other sources. For WorldCereal, the temporal observations are composited into twelve monthly timesteps. The current global coverage of WorldCereal is on the order of several million samples, including both land-cover and crop-type labels. 

For analysis we group the detailed labels into a coarser legend, for example merging winter, spring and durum wheat into a single \emph{wheat} class, or combining minor crops into groups such as fibre crops, so that each class has enough samples in a given area to define a meaningful reference population against which outliers can be judged.


\section{Embedding-based anomaly detection}
\label{sec:eo-embeddings}

\subsection{EO embeddings}

For each training sample a pretrained EO encoder produces a fixed-size vector representation of its local multi-temporal environment. 
We compute here embeddings for outlier detection through a Presto-style model~\cite{tseng2023presto}, which is very lightweight and suitable for this kind of analysis. To initially not have leakage of labels into the embeddings, we rely on the official pretrained weights to compute them. We align the embedding computation around the valid time when the crop was observed to be on the ground \cite{moletto2026enhancing}.
The EBA detector itself treats them as just $d$-dimensional vectors (here $d{=}128$) and does not depend on the architecture, requiring only that samples with similar crop types and observation conditions map to nearby vectors.

\subsection{Local slices in embedding space}

To avoid global assumptions about crop signatures, we operate on spatially local, label-consistent \emph{slices}. A slice is the set of samples sharing a cell of an H3 grid~\cite{uberh3} at a chosen level $L$ (for example $L\in\{1,2,3\}$), a label in the chosen label domain (either croptype or landcover), and optionally further grouping attributes such as year or source dataset. 
The data are grouped by the keys 
\[
\text{slice key} = (\text{h3 cell},\ \text{label}) \quad [\,+\ \text{dataset id},\ \text{year}\,].
\]
Slices with fewer than a configurable minimum number of samples (for example $50$) are merged with neighbours up to a certain range, and if still undersized they are excluded from scoring, since a stable reference cannot be defined from too few points.

For a slice with $N$ samples and embeddings $E \in \mathbb{R}^{N \times d}$ we compute, for each sample $i$, its cosine distance to the slice centroid $c$,
\[
d^{\text{centroid}}_i = 1 - \cos\!\big(\theta(E_i, c)\big),
\]
and the mean cosine distance to its $k$ nearest neighbours within the slice. The two terms are complementary: the centroid distance measures how far a point lies from the bulk of its declared crop-type or land-cover class in that slice, while the neighbour distance catches points that sit apart even from nearby samples of the same class. For moderate slice sizes we compute both exactly from the full pairwise distance matrix and for large slices we use an approximate nearest-neighbour index to avoid quadratic cost.

\subsection{Score normalization and combination}
Raw distances are mapped to approximate $[0,1]$ anomaly scores using slice-specific empirical percentiles. For a distance vector $d$ within a slice, with $p_{\min},p_{\max}$ its $q_{\min}$ and $q_{\max}$ percentiles,
\[
s_i = \text{clip}\!\left( \frac{d_i - p_{\min}}{p_{\max} - p_{\min} + \varepsilon},\ 0,\ 1 \right),
\]
applied separately to the centroid and neighbour distances and averaged into a single score $S_i = \tfrac12 s^{\text{centroid}}_i + \tfrac12 s^{\text{kNN}}_i$. From $S_i$ we additionally derive a rank-based score $S^{\text{rank}}_i$, obtained by mapping the within-slice order of $S_i$ to $[0,1]$, and a robust z-score $S^{\text{z}}_i$ based on the slice median and median absolute deviation (MAD). These capture complementary views of how unusual a sample is, so we average them into a single $\mathrm{mean\_score}$:
\[
\mathrm{mean\_score}_i = \tfrac{1}{3}\big(S_i + S^{\text{rank}}_i + S^{\text{z}}_i\big).
\]
All three terms share the same $[0,1]$ scale: the percentile and rank terms are bounded by construction, and the robust (MAD-based) z-term is mapped into $[0,1]$ by a logistic squashing of the per-slice z-scores. The $\mathrm{mean\_score}$ therefore also lies in $[0,1]$, and the equal weighting keeps any single statistic from dominating.

\subsection{Flagging and grading outliers}
\label{sec:grading}
Within each slice we flag a sample as anomalous when its score sits far above the slice's typical value. We use the median absolute deviation (MAD), a standard robust outlier rule~\cite{leys2013mad}. A sample is flagged when its score exceeds the slice median by more than $k$ MADs, with $k{=}4$, which keeps flags rare in clean slices. We additionally cap the flagged share of any slice at $10\%$ so that a few heavy-tailed slices cannot dominate. Simpler percentile or fixed-threshold rules are possible, but the MAD rule adapts to each slice's own spread and is itself robust to the outliers it must detect, which is why we prefer it.

Rather than a single yes/no decision, we grade each sample by how strongly it stands out, so that downstream training can choose how aggressively to clean:

\begin{itemize}
  \item \textbf{normal:} not flagged; treated as reliable.
  \item \textbf{flagged:} above the MAD threshold; a mild anomaly.
  \item \textbf{suspect:} flagged and among the  extreme samples (e.g.\ $S^{\text{rank}}_i$ > $0.98$).
  \item \textbf{candidate:} the most extreme of all (e.g.\ $S^{\text{rank}}_i > 0.99$), the most confident errors.
\end{itemize}
The cleaning experiments use these grades as options: candidates are the first to remove, while extending removal to suspects and flagged points discards progressively more.

\begin{figure}[t]
  \centering
  \includegraphics[width=0.62\linewidth]{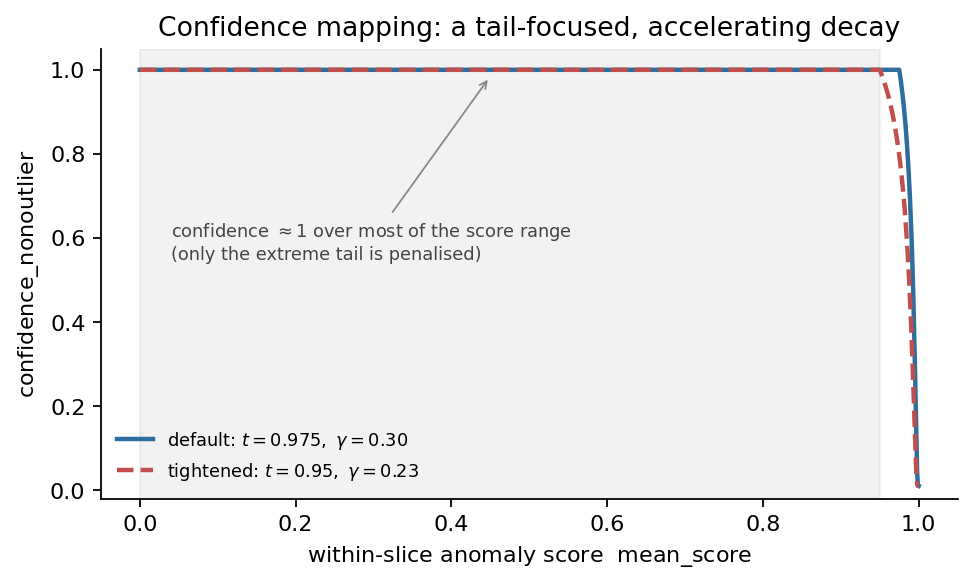}
  \caption{Mapping from the within-slice anomaly score to
  $\textit{confidence\_nonoutlier}$, for the default and tightened operating
  points; confidence stays at $1$ over most of the score range and collapses only
  in the extreme tail.}
  \label{fig:confcurve}
\end{figure}

\subsection{From mean score to confidence}
\label{sec:confidence}
The $\mathrm{mean\_score}$ lies in $[0,1]$, but only its upper end reliably separates genuine outliers, while most of the lower range reflects ordinary within-class variation. To turn it into a usable soft weight for downweighting we therefore generate $\textit{confidence\_nonoutlier}$, which stays near $1$ over the bulk of the range of mean\_score and falls only for the highest scores. It is obtained from $\mathrm{mean\_score}$ by an accelerating decay with a knee at $t$: writing $y=\mathrm{clip}\big((\mathrm{mean\_score}-t)/(1-t),\,0,\,1\big)$,
\[
\textit{confidence\_nonoutlier} = c_{\min} + (1-c_{\min})\,
\exp\!\Big(-\gamma\,\tfrac{y}{1-y}\Big),
\]
with a floor $c_{\min}$. A point scoring below $t$ keeps confidence $1$, and only the extreme tail is penalised; $\gamma$ controls how sharply confidence collapses as the score approaches $1$ (\cref{fig:confcurve}).

We report two operating points: a \textbf{default} setting ($t{=}0.975,\ \gamma{=}0.30$) and a \textbf{tightened} one ($t{=}0.95,\ \gamma{=}0.23$) that starts penalising slightly earlier and harder. Finally, we apply this down-weighting only to points the slice threshold already flagged, all unflagged points keep confidence $1$. 


\subsection{Robustness Measures}
\label{sec:robustness}
\subsubsection{Robust reference centroid:}
\label{sec:robust-centroid}
The centroid $c$ is the reference against which anomalies are measured, and computing it as the plain mean is fragile. The very outliers we wish to detect pull the mean toward themselves, deflating their own distance and \emph{masking} them, the classical masking/swamping problem in robust statistics~\cite{rousseeuw1987robust,leys2013mad}. 
We therefore use a contamination-resistant trimmed centroid, starting from the mean, we measure the cosine distance of every point, discard the farthest fraction $\tau$ (set to at least the maximum expected outlier fraction), recompute the mean on the retained inlier core, and iterate to convergence,
\[
c^{(t+1)} = \operatorname{mean}\Big(\{E_i : d^{\text{centroid}}_i(c^{(t)}) \le Q_{1-\tau}\}\Big),
\]
where $Q_{1-\tau}$ is the $(1-\tau)$ quantile of the distances to $c^{(t)}$.

\subsubsection{Down-weighting flags in uninformative neighbourhoods:}
\label{sec:slice-trust}
A distance-to-reference score is only meaningful where the embedding actually separates the relevant classes. With a frozen encoder this can fail in areas whose representation does not distinguish the crops in question, producing flags that reflect a weak embedding rather than a genuine label error. We therefore estimate, for each neighbourhood, how well its classes are separated, using a silhouette-style ratio of between-class to within-class distance
(and a simple concentration measure where only one class is present). Where this separability is very low we pull the confidence of any flags there back to $1$. In practice the gate triggers rarely, it is a safeguard against degenerate neighbourhoods.

\subsubsection{Consistency of the confidence and flagged outputs:}
\label{sec:conf-consistency}
The EBA detector gives both a discrete grading \cref{sec:grading} and the continuous $\textit{confidence\_nonoutlier}\in[0,1]$ \cref{sec:confidence}, where a value near $1$ means the sample is confidently a clean, non-outlier example and a value near $0$ means a likely error. Because the underlying score is a within-slice rank, the single highest-scoring sample of every slice, including a perfectly clean one, receives a high anomaly score and would therefore be given a low confidence purely for being its slice's relative maximum. To prevent this, we tie the continuous confidence to the discrete flag, therefore any sample in the normal grading category keeps confidence $1$, and the decay adjusts only the confidence of samples that were already flagged.



\vspace{\baselineskip} 

The whole pipeline is openly available at \url{https://github.com/WorldCereal/EBA_detector}

\providecommand{\placeholder}{\textit{TBD}}

\section{Experimental design and results}
\label{sec:experiments}

The evaluation addresses three questions: whether the flagged points behave like genuine label errors, whether acting on them improves a downstream classifier, and whether the benefit survives into the full-scale WorldCereal crop-type mapping pipeline, applied as a two-task model finetuning across five macro-regions.

We do tests against synthetic corruptions in the labels and see how accurately the detector can flag those corruptions. We also do a sweep running a fast gradient-boosted proxy catboost model on different variations of outlier treatments, and finally we run the full dual-head WorldCereal finetuning trainer on data points across five macro-regions.
Outlier treatments are applied to the training split only, the held-out split is fixed across all conditions and is split spatially ( H3 cell and source dataset) to prevent leakage. Downstream metrics are reported on
three views of that fixed split: \textbf{full} (all points kept),
\textbf{cand.\ removed} (detector-flagged candidates deleted) and
\textbf{conf.\ weighted} (each point weighted by its $\textit{confidence\_nonoutlier}$).
We sweep five regions $\times$ eleven training treatments $\times$ two detector
operating points (a \emph{default} and a \emph{tightened}-threshold
configuration),whereas for worldcereal runs we use a single seed, so from repeated baselines we estimate a run-to-run noise floor of $\approx\!0.8$ macro-F1 points and emphasize effects consistent in sign across regions. Because the best of eleven treatments is reported per region, this selection can flatter individual gains, so effects near this floor should be read as suggestive rather than established.

\subsection{Evidence that the flagged points are label errors}
\label{sec:exp-recovery}

\paragraph{Test 1: controlled synthetic ground truth.}
Because the real error process is unobserved, we create ground truth by
\emph{injecting} label noise into otherwise-trusted labels and measuring how
well the detector recovers it, re-running the full per-slice pipeline on the
corrupted labels. Three corruption models mirror real WorldCereal error modes:
\textbf{gross} (relabel to a random present class), \textbf{subtle} (relabel to
a class that co-occurs in the same H3 cell, a confusable-class error), and
\textbf{whole-dataset} (corrupt every point within the same source dataset, mimicking
a dataset digitised against the wrong legend). A purely local detector cannot, by construction, flag whole-dataset corruption, since a fully relabelled source forms its own internally consistent reference. Crucially, the detector is a
\emph{per-slice, high-precision} decision: we therefore report the
\emph{enrichment} of planted errors in the flagged set (how many times above the base rate they are concentrated), and detection Area under the ROC curve (AUROC), all on the population the detector actually scores (slices large enough to define a centroid).

\begin{figure}[t]
  \centering
  \includegraphics[width=0.78\linewidth]{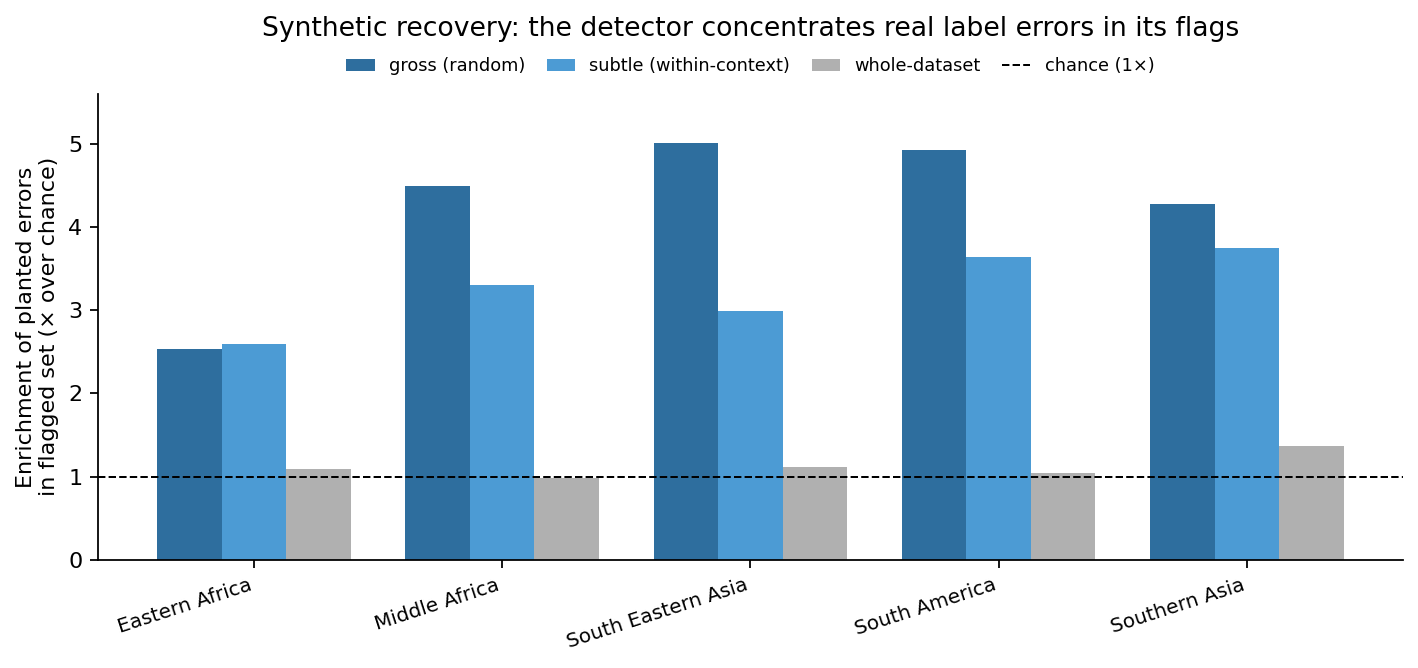}
  \caption{Enrichment of planted label errors in the flagged set, by region and
  corruption mode. Scattered errors are concentrated $2.5$--$5\times$ above chance;
  whole-dataset corruption sits at chance.}
  \label{fig:enrichment}
\end{figure}

\begin{figure}[t]
  \centering
  \includegraphics[width=0.62\linewidth]{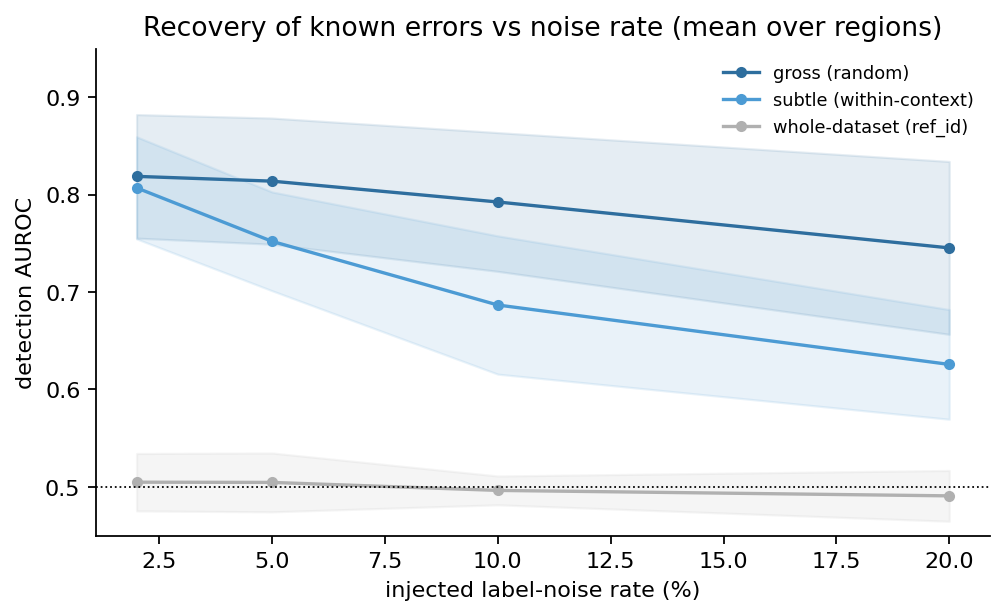}
  \caption{Detection AUROC vs.\ injected noise rate (mean over regions; band $\pm1$
  s.d.). Recovery is strongest at the low rates typical of real data; whole-dataset
  corruption stays at chance.}
  \label{fig:auroc-rate}
\end{figure}

The flagged set is strongly enriched with the planted errors
(\cref{tab:recovery}, \cref{fig:enrichment}). Gross errors are concentrated
$4.3$--$5.0\times$ above chance in four of five regions (top-$5\%$ lift up to $9\times$ with AUROC up to $0.84$), and subtle confusable-class errors, the harder and more realistic case, are still enriched $3.0$--$3.8\times$ (AUROC $\approx0.72$).
Detection is strongest at the low noise rates that dominate real reference data and degrades gracefully as noise grows (\cref{fig:auroc-rate}). This is as the embedding centroid gets affected too heavily, or the noise clusters grow too big so the detected "outliers" are no longer planted "outliers". Eastern Africa is the weakest region ($2.5\times$, AUROC $0.66$), even there the flags are $2.5\times$ enriched. Possible reasons Eastern Africa may relate to smallholder farming and mixed cropping leading to a reduced chance of identifying noise and also possibly due to limited separability of the frozen embeddings there. The case of whole-dataset corruption is the one exception where per-slice detection is at chance ($\approx\!1.0\times$, AUROC $\approx0.50$), because a fully corrupted dataset may even corrupt the centroid of its own slices and so cannot look anomalous relative to it. 

\begin{table}[t]
\centering
\small
\begin{tabular}{@{}lcccc@{}}
\toprule
& \multicolumn{2}{c}{Gross} & \multicolumn{2}{c}{Subtle} \\
\cmidrule(lr){2-3}\cmidrule(lr){4-5}
Region & AUROC & Enrich.$\times$ & AUROC & Enrich.$\times$ \\
\midrule
Eastern Africa     & $0.66_{\pm.00}$ & $2.5_{\pm.2}$ & $0.67_{\pm.01}$ & $2.6_{\pm.1}$ \\
Middle Africa      & $0.83_{\pm.01}$ & $4.5_{\pm.1}$ & $0.75_{\pm.02}$ & $3.3_{\pm.2}$ \\
South-Eastern Asia & $0.84_{\pm.02}$ & $5.0_{\pm.5}$ & $0.66_{\pm.01}$ & $3.0_{\pm.1}$ \\
South America      & $0.83_{\pm.01}$ & $4.9_{\pm.1}$ & $0.76_{\pm.00}$ & $3.6_{\pm.0}$ \\
Southern Asia      & $0.79_{\pm.00}$ & $4.3_{\pm.1}$ & $0.75_{\pm.00}$ & $3.8_{\pm.1}$ \\
\bottomrule
\end{tabular}
\caption{Synthetic-noise recovery (crop type). Enrichment is the concentration of planted errors in the flagged set relative to chance. Values are mean$_{\pm\text{std}}$ over three seeds and four injected-noise rates, indicating the recovery is stable across seeds. Whole-dataset corruption case is omitted as it
sits at chance ($\approx\!1.0\times$, AUROC $\approx0.50$).}
\vspace{-10pt}
\label{tab:recovery}
\end{table}

\begin{figure}[h]
  \centering
  \includegraphics[width=0.9\linewidth]{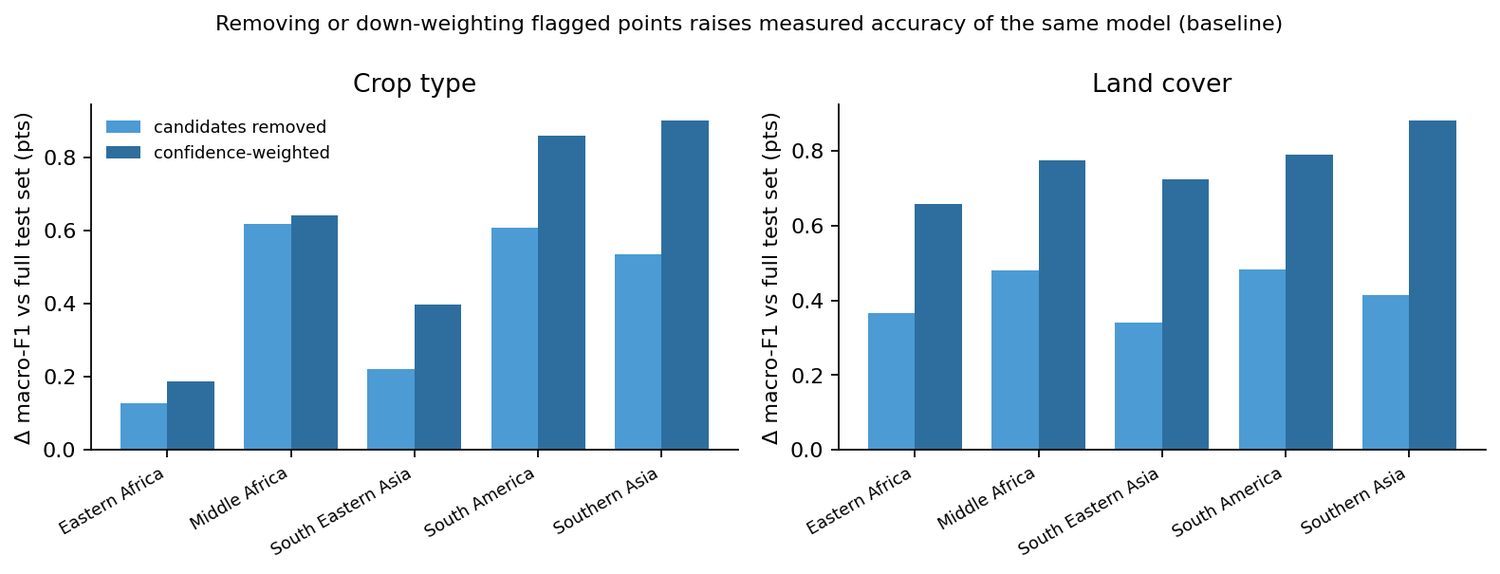}
  \caption{Cleaning the \emph{test} labels of a fixed (baseline) model raises its
  measured macro-F1 in every region, for both tasks.}
  \vspace{-10pt}
  \label{fig:testview}
\end{figure}

\paragraph{Test 2: model-independent test on real data.}
The same conclusion holds without any synthetic noise. For \emph{any} fixed
trained model, we re-score its predictions on the held-out set after removing the detector-flagged points or down-weighting them by confidence. Because the model is unchanged and only the treatment of the test labels differs, any rise in the measured score indicates that the flagged points are disproportionately unreliable for the model, whether through mislabelling, misplacement, or being genuinely hard to classify. This test on its own cannot separate ``mislabelled'' from ``correct but hard'', and because the downstream model shares the Presto representation used for detection, part of the effect may reflect points that are simply hard in that embedding space. The synthetic recovery above, built from known injected errors, is what ties the flags to actual mislabelling; the two tests are complementary.
The effect is essentially deterministic (\cref{tab:testview},
\cref{fig:testview}): across the $55$ trained models per detector configuration,
down-weighting the flagged points raises macro-F1, weighted-F1 and accuracy in
$54/55$ models for crop type and $55/55$ for land cover. It holds for both detector configurations, so the conclusion does not depend on threshold. 

We cannot fully rule out that a valid observation which is fully out of distribution (>MAD 4.0) may be marked as an outlier as well. Some evidence that point to actual mislabelled identifications to counter this are the visualized surfaced points (\cref{fig:out-examples}), the recovery of synthetic mislabelled points (\cref{fig:auroc-rate}) and the ablation with the finetuned encoder (\cref{fig:encoder}). However, some iterations still may be needed to identify true noise, that doesn't include the valid signal range around the centroids too much. 

\begin{table}[t]
\centering
\small
\begin{tabular}{@{}llccc@{}}
\toprule
Detector & Task & $\Delta$macro-F1 & $\Delta$wtd-F1 & $\Delta$acc. \\
\midrule
tightened & Crop type  & $+0.59$ (54/55) & $+0.30$ (55/55) & $+0.30$ (55/55) \\
tightened & Land cover & $+0.81$ (55/55) & $+0.65$ (55/55) & $+0.66$ (55/55) \\
default   & Crop type  & $+0.39$ (54/55) & $+0.20$ (55/55) & $+0.21$ (55/55) \\
default   & Land cover & $+0.54$ (55/55) & $+0.42$ (55/55) & $+0.43$ (55/55) \\
\bottomrule
\end{tabular}
\caption{Model-independent test-view lift. For a fixed trained model,
down-weighting the flagged held-out points by confidence changes the measured
score by the mean shown (points), with the number of models improved out of $55$
($5$ regions $\times$ $11$ treatments).}
\vspace{-10pt}
\label{tab:testview}
\end{table}

\subsection{Cleaning the training labels improves downstream models}
\label{sec:exp-downstream}

\begin{figure}[t]
  \centering
  \includegraphics[width=0.82\linewidth]{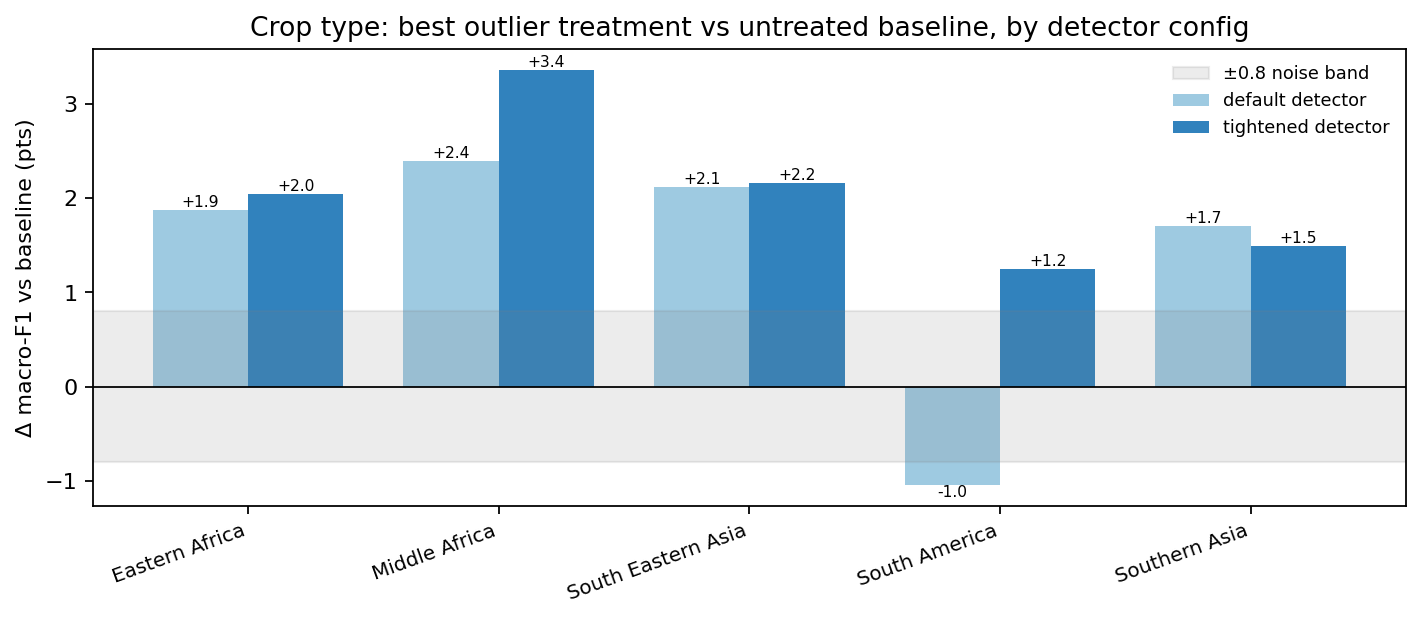}
  \caption{Crop-type model improvement of the best treatment over baseline, per region/detector configuration. Shaded band is $\pm0.8$-point noise floor. The tightened detector is positive in all five regions.}
  \label{fig:bestdelta}
  \vspace{-20pt}
\end{figure}

\paragraph{WorldCereal regional model trainings}
We train the dual-head (land cover + crop type) WorldCereal model under each treatment across all five regions. With the tightened detector \cref{sec:confidence} the best outlier treatment improves crop-type macro-F1 in \emph{all five} regions ($+1.2$ to $+3.4$, mean $+2.1$) and land cover in all five ($+0.3$ to $+1.7$); with the default detector four of five improve (\cref{tab:regions}, \cref{fig:bestdelta}).
The benefit is realised by \emph{conservative} treatment: across regions the strongest policies are candidate removal and confidence/quality weighting, whereas aggressively dropping \emph{all} flagged points is the worst policy on average 
discarding hard-but-sometimes-correct boundary samples.
The one configuration that leads to worse performance is South America ($-1.0$ crop type), a region with seasonality and legend-alignment issues, tightening the detector parameters still recovers it ($+1.2$).The best outlier treatment in the regional training pipelines is region dependent: for some cleaner regions only dropping candidate outlier \cref{sec:grading} shows best performance,  whereas for noisier regions, stronger filtering proves optimal.

\begin{table}[t]
\centering
\begin{minipage}[b]{0.55\linewidth}
\centering\small
\setlength{\tabcolsep}{3pt}
\begin{tabular}{@{}llccc@{}}
\toprule
Region & Task & base & best & $\Delta$ \\
\midrule
Eastern Africa     & CT & 0.342 & 0.362 & \textbf{+2.0} \\
Middle Africa      & CT & 0.786 & 0.820 & \textbf{+3.4} \\
South-Eastern Asia & CT & 0.905 & 0.926 & \textbf{+2.2} \\
South America      & CT & 0.584 & 0.596 & +1.2 \\
Southern Asia      & CT & 0.714 & 0.729 & \textbf{+1.5} \\
\midrule
Eastern Africa     & LC & 0.592 & 0.602 & +1.0 \\
Middle Africa      & LC & 0.706 & 0.719 & \textbf{+1.3} \\
South-Eastern Asia & LC & 0.804 & 0.821 & \textbf{+1.7} \\
South America      & LC & 0.753 & 0.756 & +0.4 \\
Southern Asia      & LC & 0.675 & 0.678 & +0.3 \\
\bottomrule
\end{tabular}
\caption{Macro-F1 (full test view, tightened detector): best treatment vs.\ untreated baseline; all ten region$\times$task cells improve.}
\label{tab:regions}
\end{minipage}\hfill
\begin{minipage}[b]{0.42\linewidth}
\centering
\includegraphics[width=\linewidth]{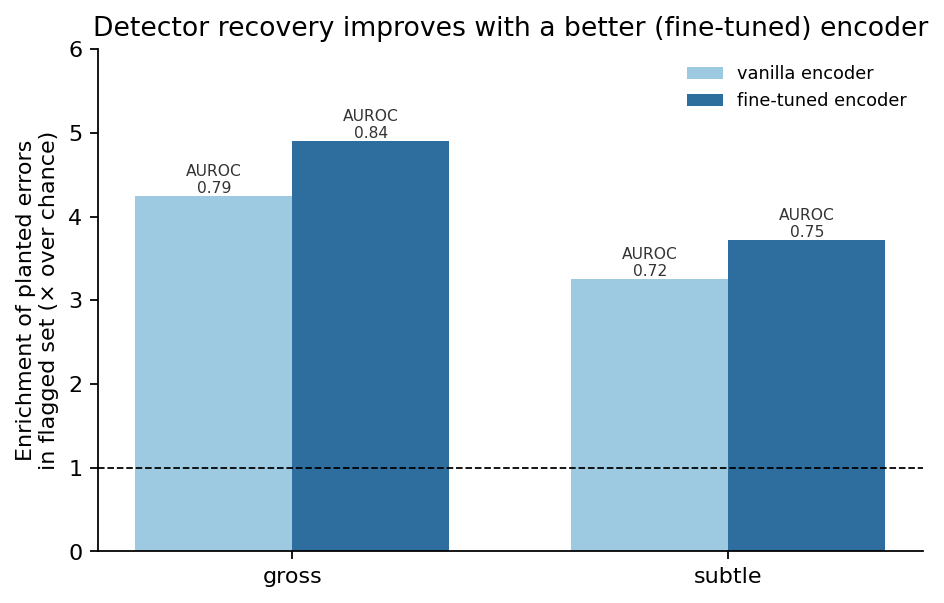}
\captionof{figure}{Synthetic-noise recovery with the vanilla vs.\ fine-tuned WorldCereal encoder (crop type, mean over regions): a stronger encoder concentrates planted errors more strongly in the flagged set, for both gross and subtle noise.}
\label{fig:encoder}
\end{minipage}
\end{table}

\paragraph{Fast proxy.}
To iterate through the different treatment options and with multiple seed values so that we get an estimate of the noise, we trained a gradient-boosted classifier (CatBoost) on the frozen embeddings, varying only the training data per \cref{tab:scenarios} (three seeds). Across the regions available, the most reliable treatment was confidence-thresholded filtering (\texttt{filter\_0.90}, mean $+0.7$ macro-F1), and aggressive removal of all flagged points is again the worst ($-1.1$), echoing the full worldcereal training outcome of ``don't over-clean'' signal. Macro-F1 over many rare crops was noisy at three seeds.

\begin{table}[t]
\centering
\small
\begin{tabular}{@{}lll@{}}
\toprule
Scenario & Train treatment & Notes \\
\midrule
\texttt{baseline}        & none & reference \\
\texttt{drop\_candidate} & remove candidates & conservative removal \\
\texttt{drop\_suspect}   & remove cand.+susp. & \\
\texttt{drop\_flagged}   & remove all flagged & aggressive removal \\
\texttt{downweight\_*}   & $w=\mathrm{conf}^{1,2,4}$ & keep all, soft weight \\
\texttt{filter\_0.90/0.95} & drop conf.\,$<\tau$ & threshold filter \\
\bottomrule
\end{tabular}
\caption{Outlier-treatment scenarios applied to the training split.}
\vspace{-10pt}
\label{tab:scenarios}
\end{table}

\subsection{Where the gains come from}
\label{sec:exp-perclass}

\begin{figure}[t]
  \centering
  \includegraphics[width=0.9\linewidth]{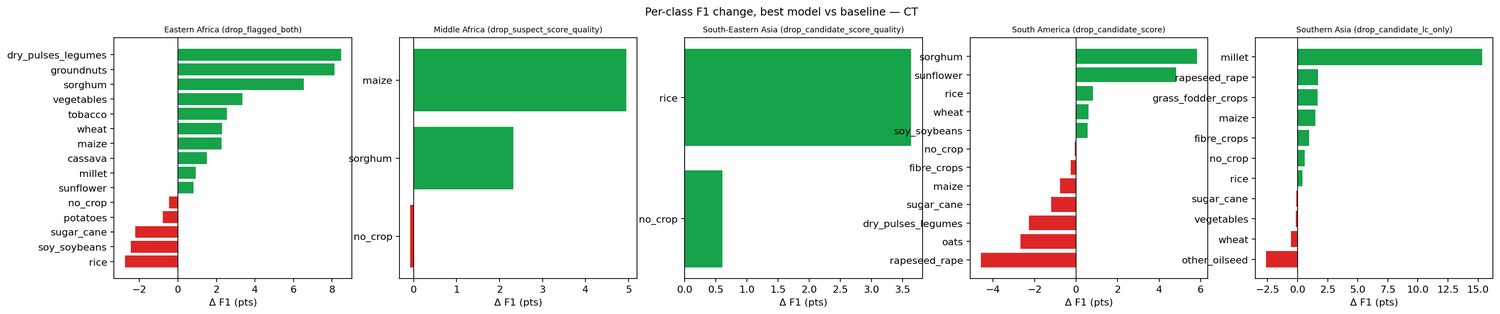}
  \caption{Per-class crop-type F1 change (best model vs.\ baseline) by region.
  Gains concentrate on rare/confusable crops; majority crops are unaffected.}
  \vspace{-3pt}
  \label{fig:perclass}
\end{figure}

\begin{figure}[t]
  \centering
  \includegraphics[width=0.75\linewidth]{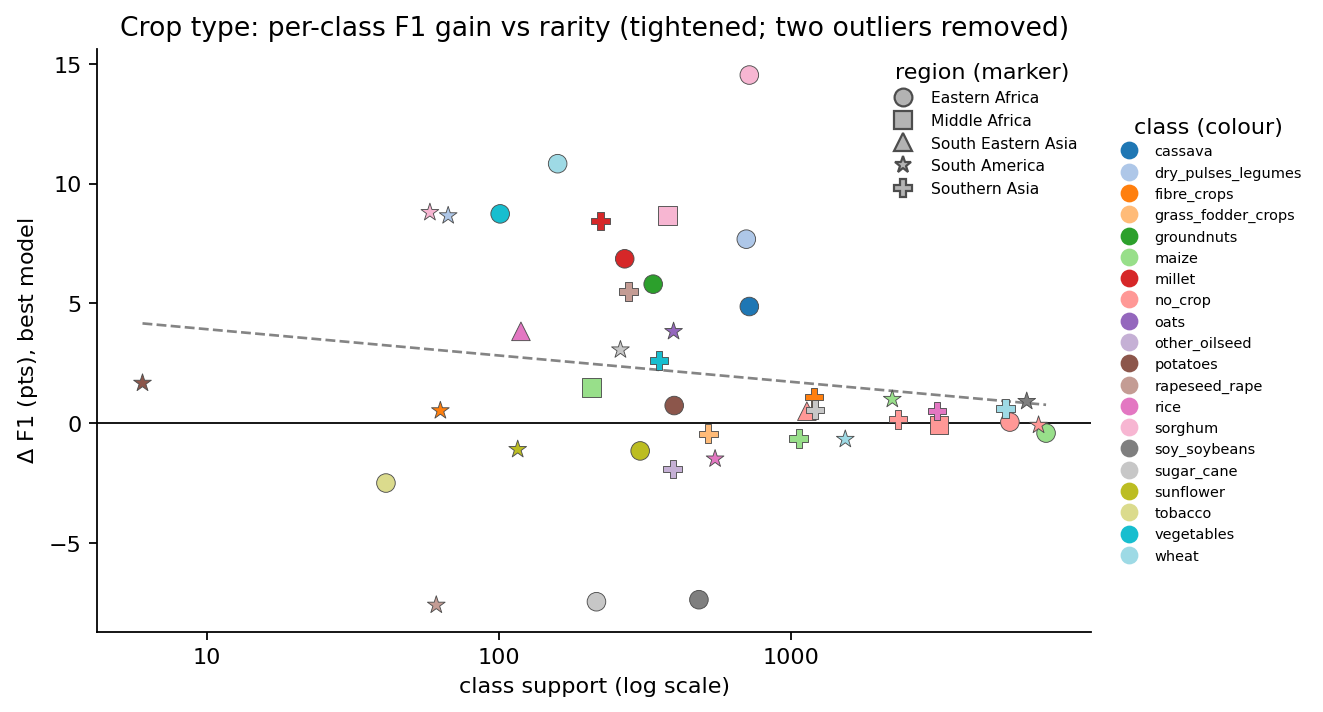}
  \caption{Per-class crop-type F1 gain vs.\ class support}
  \vspace{-12pt}
  \label{fig:rarity}
\end{figure}

\begin{figure}[t]
  \centering
  \includegraphics[width=0.72\linewidth]{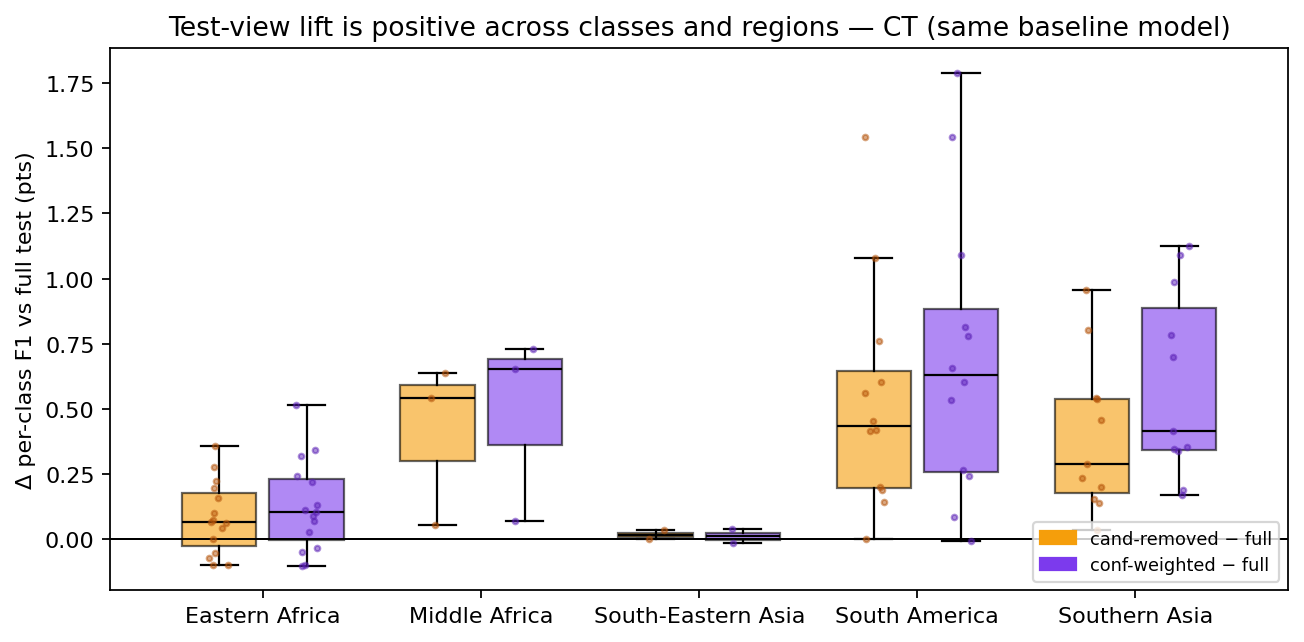}
  \caption{Per-class test-view lift for a fixed baseline model
  (deltas +ve from the full view).}
  \label{fig:classwise-lift}
  \vspace{-10pt}
\end{figure}

Because macro-F1 weights all classes equally, an improvement is informative only
if it comes from the hard classes, and it does (\cref{fig:perclass}). The
crop-type gain is carried almost entirely by rare, confusable crops while
well-populated majority crops are unchanged: in Southern Asia \emph{millet}
improves by $+15$ F1 points, with further gains on oilseeds, fodder grasses and
minor cereals elsewhere; smaller classes gain the most (\cref{fig:rarity}). The
per-class test-view lift is likewise positive across the great majority of
classes and regions (\cref{fig:classwise-lift}).

\subsection{Ablations}
\label{sec:exp-ablation}

\paragraph{Detector operating point.}
Running the whole grid under both the default and the tightened confidence mapping (\cref{fig:confcurve}) leaves the conclusions unchanged in sign: the
model-independent test-view lift is present under both and slightly larger under the tightened detector, which also gives the stronger result
(best-treatment mean $+2.1$ vs.\ $+1.4$ crop-type macro-F1; positive in $5/5$
vs.\ $4/5$ regions).

\vspace{-5pt}
\paragraph{Encoder.}
The detector treats the embedding as an opaque vector, so a stronger encoder should expose label errors more sharply. Re-running the synthetic-recovery test on embeddings from the global \emph{fine-tuned} WorldCereal encoder rather than the vanilla one improves synthetic error recovery rates: gross-error enrichment rises from $4.3\times$ to $4.9\times$ (AUROC $0.79\!\to\!0.84$) and subtle-error enrichment from $3.3\times$ to $3.7\times$ ($0.72\!\to\!0.75$, \cref{fig:encoder}). The downstream signal persists, with the best treatment improving crop-type macro-F1 in every available region under the fine-tuned
encoder, including South America ($+2.0$), and the test-view lift positive in $30/30$ models.

Because the detector is agnostic to the embedding length, the same pipeline runs
unchanged on representations from other encoders; we provide tooling to score
$64$-dimensional AlphaEarth~\cite{brown2025alphaearth} embeddings, and as a qualitative example run the EBA detector on AlphaEarth embeddings of the Southern Asia region, the detector surfaces analogous misplaced and mislabelled points (\cref{fig:out-examples}, lower panel), showing again that an encoder with richer representations will surface labeling anomalies in the reference datasets using the EBA detector workflow.  
Two baseline comparisons further support the method (\cref{tab:baselines}): dropping the H3
locality lowers gross recovery from $0.79$ to $0.66$ AUROC, and off-the-shelf Isolation
Forest / Local Outlier Factor on the same embeddings sit at chance.
\vspace{-10pt}
\section{Discussion}
\label{sec:discussion}

The embedding-based approach scores anomalies on a representation that already integrates temporal, spectral and contextual information into a compact vector, so detection is decoupled from low-level preprocessing choices such as interpolation and modality scaling. Locality via H3 slicing ensures that each sample is judged against nearby, label-consistent neighbours rather than against global statistics dominated by dense regions and majority crops.  

A key consideration is that embeddings generally come from a frozen, un-fine-tuned encoder, so outlier scores are only as good as the separation that this generic representation provides for a given region and class. The slice-trust gate checks that a slice's embedding geometry is informative, but it cannot create separation the encoder lacks. This suggests an upside in choosing a highly discriminative encoder for the target application, in which a cleaner training set yields a better fine-tuned encoder whose embeddings in turn support better accuracies (\cref{sec:exp-downstream} and \cref{sec:exp-ablation}).

The robustness
mechanisms of \cref{sec:robust-centroid}
target the specific ways this can fail in practice: the trimmed centroid removes
self-masking by outliers; the safeguard against
degenerate neighbourhoods \cref{sec:slice-trust} prevents an uninformative
embedding geometry from manufacturing flags; group-level aggregation surfaces
whole datasets that are systematically off but that point-wise neighbours hide;
and only flagged samples receive confidence-based soft weighting.

Several limitations remain. In regions or classes with very few samples per slice, statistical anomaly detection is inherently unreliable; the minimum slice size, resolution-aware merging and neutral scoring mitigate but cannot eliminate this. Embeddings inherit any systematic bias of the pretrained model, which propagates into the scores. The real-data test of \cref{sec:exp-recovery} shows the flagged points are disproportionately hard for a fixed model but does not by itself prove mislabelling, and the operational WorldCereal runs use a single seed, so region-level gains near the noise floor are best read as suggestive. Some flagged points may correspond to legitimate but
rare management practices or mixed pixels rather than hard label errors, so the choice of which categories to remove versus down-weight is
a precision/recall trade-off best set per application, which is why we report the
full removal-to-weighting spectrum rather than a single operating point. Finally,
as \cref{sec:exp-recovery} shows, the point-wise detector is near-chance against
whole-dataset corruption by construction.
While we run the detector on AlphaEarth embeddings as a qualitative check, a full downstream benchmark across other geospatial foundation models is left to future work. The framework extends naturally to active learning, in which candidate anomalies
are routed to experts whose feedback refines thresholds, to complementing the
distance and neighbour score with off-the-shelf detectors such as isolation
forests~\cite{liu2008isolation} or the local outlier factor~\cite{breunig2000lof}
applied to the same embeddings; and to explicit label-noise modelling in the
downstream model. 

\vspace{-12pt}
\section{Conclusion}
\label{sec:conclusion}

We presented a locality-aware, embedding-based framework for detecting and
treating outlier samples in large-scale Earth-observation reference datasets,
with crop-type and landcover training data as the target application. Beyond the core
slice-wise scoring, we contributed robustness mechanisms that close concrete
gaps in embedding-space detection, a contamination-resistant centroid, a
gating that accounts for the potential collapse of the un-fine-tuned encoder in a region, group-level
aggregation for systematically mislabelled datasets. Two complementary tests indicate that the flagged points behave like genuine label errors: recovery of injected errors against synthetic ground truth, and a model-independent test on real data (which shows they are disproportionately unreliable for a fixed model, though it cannot alone separate mislabelled from merely hard cases), and a downstream evaluation on a fast proxy
and the WorldCereal style model shows that conservative cleaning improves
crop-type accuracy across regions while over-cleaning hurts. The result is a
pipeline for embedding-space based label cleaning, and a
clear path toward strengthening it
further.

\subsubsection*{Acknowledgements.}
This work was carried out within the WorldCereal project. The WorldCereal project is funded by the \href{https://www.esa.int/}{European Space Agency (ESA)} under grant no.\ 4000130569/20/I-NB.

\bibliographystyle{splncs04}
\bibliography{main}

\clearpage
\appendix
\section*{Supplementary Material}
\vspace{-28pt}
\setcounter{figure}{0}
\renewcommand{\thefigure}{S\arabic{figure}}
\setcounter{table}{0}
\renewcommand{\thetable}{S\arabic{table}}

\begin{figure}[H]
  \centering
  \includegraphics[height=0.87\textheight]{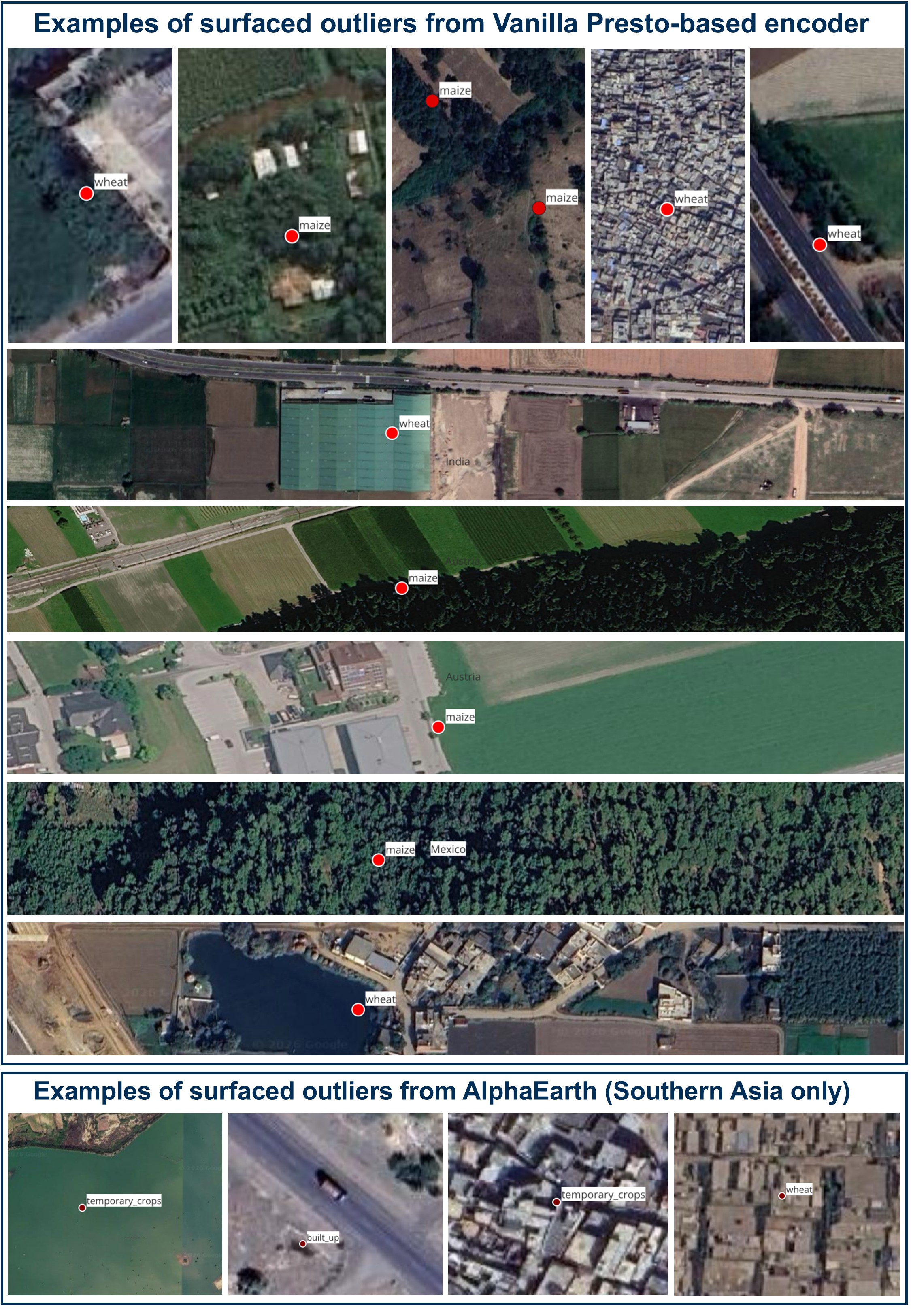}
  \vspace{-10pt}
  \caption{Examples of outliers surfaced by the EBA detector. The upper panels use  vanilla Presto encoder; the lowest panel use AlphaEarth embeddings (Southern Asia only). Some are clearly misplaced or wrongly labelled points, yet locating them among $7.3$ million samples is far from straightforward; others are subtler and not obviously odd on a basemap alone.}
  \label{fig:out-examples}
\end{figure}
\vspace{-12pt}
\subsection*{Manual visual audit of flagged points}
As a complementary real-data check, an annotator visually inspected flagged
\emph{candidate} and \emph{suspect} points in QGIS over high-resolution
satellite imagery and labelled each as \emph{good} (looks like a genuine
anomaly), \emph{wrong} (looks like a correctly-labelled point that should not
have been flagged), or \emph{uncertain} (not adjudicable from imagery, e.g.\ a
dry riverbed, or a point whose basemap date is offset from the observation).

Two patterns stand out (\cref{tab:audit}). First, for every crop-type class
(temporary crops, wheat, maize, other temporary crops) none of the decideable
outlier points were judged \emph{wrong}, for either encoder and for both candidates and
suspects; the remaining non-\emph{good} share is \emph{uncertain} rather than a
false positive. Because a visual audit cannot see the multi-temporal, multi-modal
cues the detector actually uses, these \emph{good} rates are a lower bound on
precision. Second, the crop classes that matter most for WorldCereal (cropland and temporary
crops) were checked most closely and show no clear false positives; the few
\emph{wrong} judgements fall in shrubland, a land-cover class for which a single,
temporally-offset basemap is an unreliable reference, since ambiguous sparse
vegetation can look mislabelled either way. AlphaEarth embeddings generally sharpen
agreement on the crop classes (consistent with the encoder ablation of
\cref{sec:exp-ablation}: a stronger representation yields cleaner flags. The
audit is small and imagery-limited, so we treat it as supporting rather than
definitive evidence; together with the synthetic recovery of
\cref{sec:exp-recovery} it indicates the outlier crop-type points are rarely
simple false positives.

\begin{table}[t]
\centering
\footnotesize
\setlength{\tabcolsep}{4pt}
\begin{tabular}{@{}lccccccc@{}}
\toprule
& & \multicolumn{3}{c}{Presto (\%)} & \multicolumn{3}{c}{AlphaEarth (\%)} \\
\cmidrule(lr){3-5}\cmidrule(lr){6-8}
Class & $n_{\text{P}}/n_{\text{AE}}$ & Good & Wrong & Unc. & Good & Wrong & Unc. \\
\midrule
\multicolumn{8}{@{}l}{\emph{Candidates}}\\
Temporary crops    & 30/30 & 66.7 & 0.0  & 33.3 & 73.3 & 0.0  & 26.7 \\
Wheat              & 11/24 & 81.8 & 0.0  & 18.2 & 76.9 & 0.0  & 23.1 \\
Maize              & 29/24 & 65.5 & 0.0  & 34.5 & 95.8 & 0.0  & 4.2  \\
Other temp.\ crops & 24/20 & 37.5 & 0.0  & 62.5 & 65.0 & 0.0  & 35.0 \\
Shrubland          & 13/15 & 46.2 & 30.8 & 23.1 & 40.0 & 20.0 & 40.0 \\
\midrule
\multicolumn{8}{@{}l}{\emph{Suspects}}\\
Temporary crops    & 30/30 & 50.0 & 0.0  & 50.0 & 63.3 & 0.0  & 36.7 \\
Wheat              & 17/9  & 47.1 & 0.0  & 52.9 & 77.8 & 0.0  & 22.2 \\
Maize              & 30/30 & 70.0 & 0.0  & 30.0 & 76.7 & 0.0  & 23.3 \\
Other temp.\ crops & 30/30 & 36.7 & 0.0  & 63.3 & 56.7 & 0.0  & 43.3 \\
Shrubland          & 19/16 & 47.4 & 42.1 & 10.5 & 37.5 & 37.5 & 25.0 \\
\bottomrule
\end{tabular}
\caption{Manual visual audit of flagged points (percent of audited points per
class; $n$ audited shown for Presto/AlphaEarth). \emph{Wrong} is $0\%$ for all
crop-type classes under both encoders; the few disagreements fall in shrubland, where imagery is an unreliable reference. \emph{Uncertain} points are
unverifiable from imagery, not confirmed correct
}
\label{tab:audit}
\end{table}

\vspace{-10pt}
\begin{table}[t]
\centering
\footnotesize
\setlength{\tabcolsep}{7pt}
\begin{tabular}{@{}lcccc@{}}
\toprule
& \multicolumn{2}{c}{Gross} & \multicolumn{2}{c}{Subtle} \\
\cmidrule(lr){2-3}\cmidrule(lr){4-5}
Detector & AUROC & Enrich.$\times$ & AUROC & Enrich.$\times$ \\
\midrule
EBA (locality-aware)     & $0.79_{\pm.07}$ & $4.2_{\pm.9}$ & $0.72_{\pm.05}$ & $3.3_{\pm.4}$ \\
Global (no H3 locality)  & $0.66_{\pm.07}$ & $3.3_{\pm.7}$ & $0.62_{\pm.07}$ & $2.9_{\pm.5}$ \\
Isolation Forest         & $0.50_{\pm.00}$ & $1.0_{\pm.0}$ & $0.50_{\pm.00}$ & $1.0_{\pm.0}$ \\
Local Outlier Factor     & $0.50_{\pm.00}$ & $1.0_{\pm.0}$ & $0.50_{\pm.00}$ & $1.0_{\pm.0}$ \\
\bottomrule
\end{tabular}
\caption{Baseline comparison on the synthetic-recovery benchmark (crop type),
mean$_{\pm\text{std}}$ over the five regions. Removing the H3 locality (\emph{global}:
one slice per label) drops recovery, so the local slicing contributes materially. Off-the-shelf
detectors run on the \emph{same} embeddings without the class-conditional, local
formulation (Isolation Forest, Local Outlier Factor) sit at chance
($\text{AUROC}\approx0.50$, enrichment $\approx1\times$), 
}
\label{tab:baselines}
\end{table}

\vspace{-10pt}
\begin{figure}[h]
  \centering
  \includegraphics[width=0.99\linewidth]{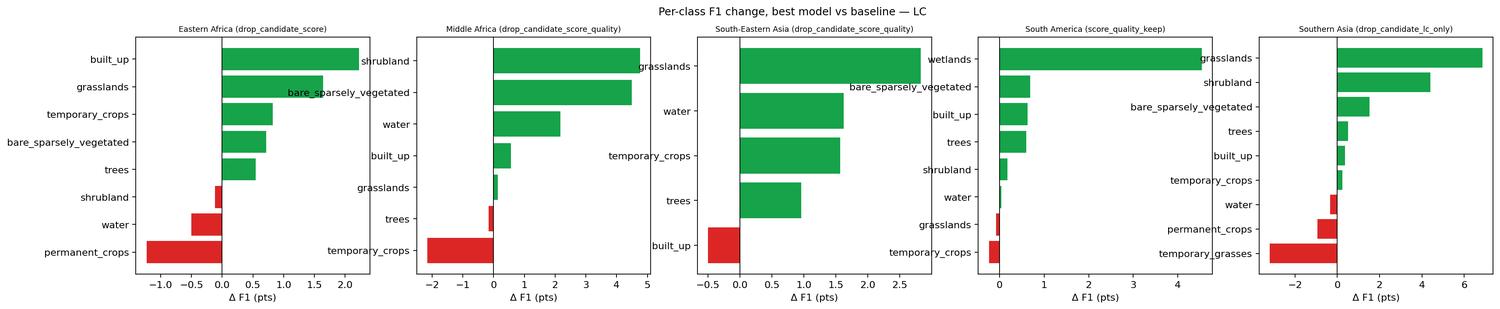}
  \caption{Per-class landcover F1 change (best model vs.\ baseline) per region}
  \label{fig:supp-lc-perclass}
\end{figure}

\vspace{10pt}
\begin{figure}[h]
  \centering
  \includegraphics[width=\linewidth]{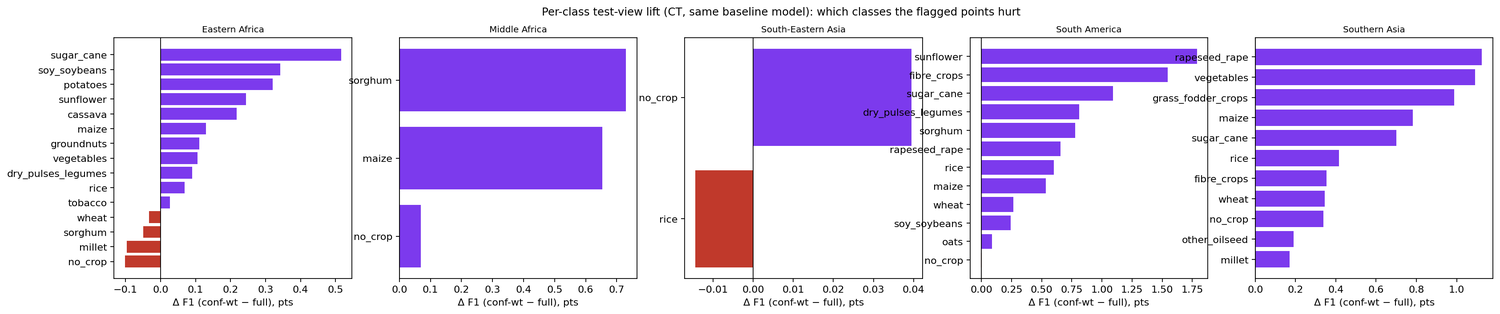}
  \caption{Per-class test-view lift (confidence-weighted minus full, same baseline
  model) per region: the classes whose held-out points the detector flags are
  those on which the model most improves once those points are down-weighted.}
  \label{fig:supp-classwise}
\end{figure}

\vspace{-15pt}
\begin{figure}[t]
  \centering
  \includegraphics[width=0.75\linewidth]{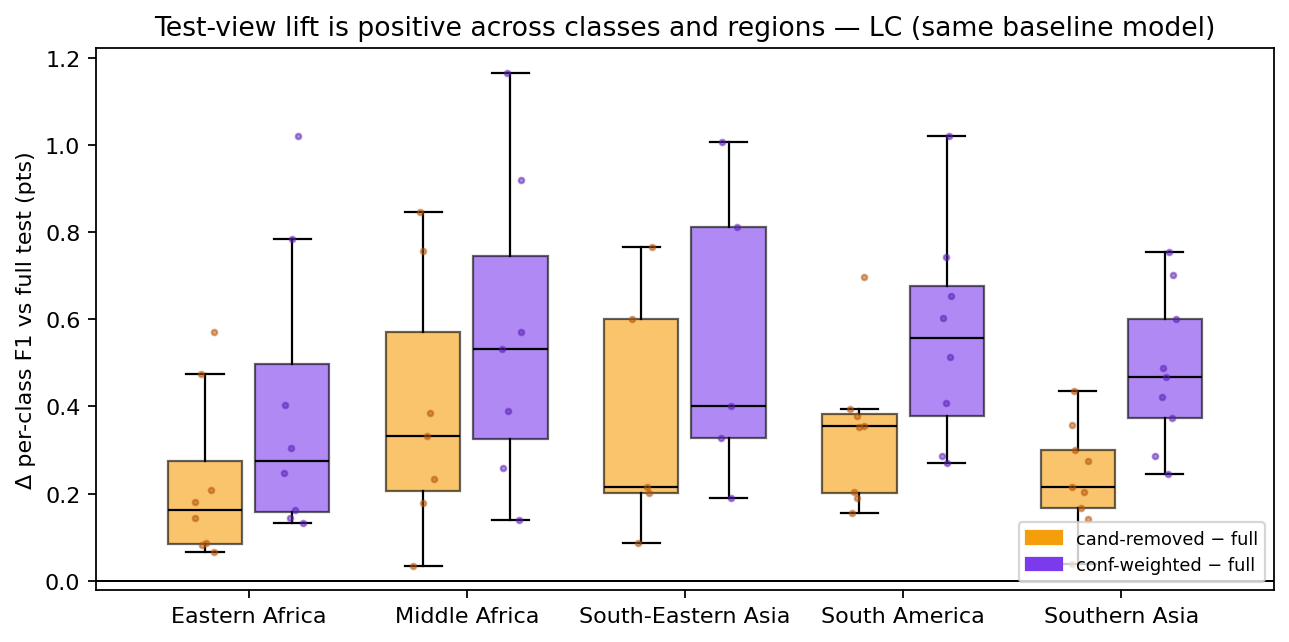}
  \caption{Landcover counterpart of the class-wise test-view lift: per-class
  F1 change (same baseline model) when flagged points are removed or
  confidence-weighted.}
  \label{fig:supp-classwise-lc}
\end{figure}

\end{document}